\documentclass[letterpaper, 10 pt, conference]{ieeeconf}

\IEEEoverridecommandlockouts                              

\usepackage{multirow}
\usepackage{graphicx} 

\usepackage{url}
\usepackage{xcolor}
\usepackage{algorithm}
\usepackage[noend]{algpseudocode} 

\usepackage{amssymb,amsmath,comment,cite}
\usepackage{booktabs}
\usepackage{tabularx}
\usepackage{threeparttable}
\usepackage{array}
\usepackage{makecell}
\usepackage{hyperref}

\usepackage{amsthm} 

\usepackage{ifthen}
\newboolean{shortver}
\setboolean{shortver}{true}

\title{\bf Linear Twisting Sidewinding Gaits for Twistable Snake Robot \\ Motion Planning on Complex Terrains}

\author{Qingling Fan$^{1}$, Zhongqiang Ren$^{1\dagger}$ 
\thanks{$^{1}$Global College, Shanghai Jiao Tong University, China (\texttt{\{qingling.fan, zhongqiang.ren\}@sjtu.edu.cn}).}
\thanks{$^\dagger$ Corresponding author}
}

\usepackage{xspace}

\usepackage{etoolbox}
\makeatletter
\patchcmd{\@makecaption}
  {\scshape}
  {}
  {}
  {}
\patchcmd{\@makecaption}
{\\}
{.\ }
{}
{}
\makeatother

\begin{document}

\maketitle

\graphicspath{{./figures/}}

\newcommand{\abbrTSnake}{T-Snake\xspace}

\begin{abstract}
    Sidewinding is an efficient gait for snake robots to locomote on various terrains, by undulating the backbone curve and shifting the contact segments along it.
However, relying solely on undulatory motion limits the locomotive capability of snake robots, and combining it with twisting motion can improve their speed and robustness.
This hypothesis has recently been verified by the rolling vector approach, which locomotes the robot like a differential-drive car.
Yet its twisting is concentrated in a short buffer segment, imposing most of the workload on the two twist joints in the middle, which can often overburden the motors and limit its locomotion.
With that in mind, this paper develops a new family of twisting sidewinding gaits that distribute the workload over the entire body with our proposed linear twisting function.
Under matched turning speed and total output power, our gaits reduce the peak output power of a single twist joint by 42\%.
Real-robot experiments show that, with the reduced workload, our new gaits can locomote the robot in various terrains that the existing rolling vector approach cannot.
Besides, our new gaits can move up to 49\% faster than conventional sidewinding without twisting.


\end{abstract}

\graphicspath{{./figure/}}
 
\section{Introduction}\label{sec:intro}

Snake robots are hyper-redundant limbless robots that can traverse complex environments~\cite{pettersen2017snake,marvi2014sidewinding,vaquero2024eels} such as sandy slopes, pipes, and compact spaces, using versatile gaits, which are cyclic motions in the shape space.
Among these gaits, sidewinding is regarded as highly efficient and robust in many terrains~\cite{tanev2005automated,astley2015modulation,ariizumi2017dynamic,malayjerdi2019analytical}.
During sidewinding, the snake keeps its body lifted except for a few static contact segments, and undulates its backbone curve both horizontally and vertically to continuously shift these segments, transporting itself using the friction at contact.

\begin{figure}[t]
    \centering
    \includegraphics[width=1.0\linewidth]{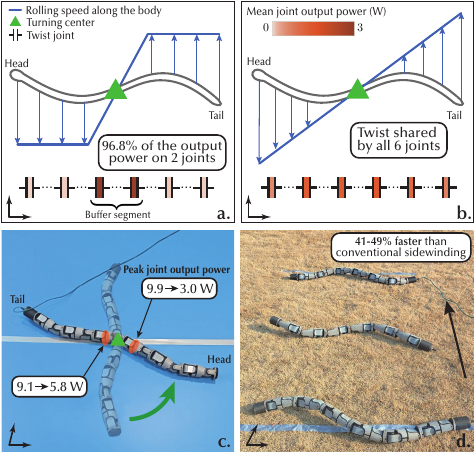}
    \caption{Twist-joint load distribution of the two gaits.
    The proposed LTS spreads the twisting over the entire body, whereas the rolling vector approach~\cite{fan2026motion} concentrates it in a narrow buffer segment.
    (a) Rolling vector: $96.8\%$ of the total output power is carried by the two twist joints inside the buffer segment.
    (b) LTS: the twisting is uniform along the body, so all six twist joints share the load; the color bar gives the mean output power of each joint.
    (c) In-place turning on the \abbrTSnake hardware: compared with the rolling vector approach, the peak output power of a single joint drops from $9.9$ to $3.0$~W at joint~3 and from $9.1$ to $5.8$~W at joint~4.
    (d) The LTS traverses four types of terrain $41$--$49\%$ faster than conventional sidewinding.}
    \label{fig:head}
\end{figure}

While being highly efficient and robust, sidewinding is still relatively slow to traverse even flat ground, and may get trapped on uneven terrains.
In contrast, the rolling motion of wheels is way faster on many complex terrains.
We take the view that relying solely on undulatory motion limits the locomotive capability of snake robots, and that combining it with rolling motion can improve both efficiency and robustness without losing the various gaits.
This hypothesis has been partially verified in simulation by the recent literature~\cite{fan2026motion}, which considers locomotion of a twistable snake robot (\abbrTSnake), a modular snake-robot mechanism that can not only bend but also twist about its backbone curve.
A rolling vector approach \cite{fan2026motion} is proposed to coordinate the motion of joints, which kinematically reduces the \abbrTSnake to a differential-drive car.
This rolling vector approach twists by assigning different rolling speeds to the head and tail segments, bridged by a short buffer segment.
As a result, the twisting is concentrated within the buffer segment, and burdens the two twist joints in the middle of the body, while leaving the others almost idle (Fig.~\ref{fig:head}(a)), which can often overload the motors in the buffer segment and limit its locomotion capability.



To coordinate the large number of joints while avoiding such uneven workload, this paper proposes a new linear twisting approach that distributes the twisting motion over the entire body.
Specifically, we let the twisting vary linearly along the backbone curve, and refer to the resulting gaits as \textit{linear twisting sidewinding} (LTS).
Following the backbone curve approach~\cite{chirikjian2002modal,yamada2006study}, this linear twisting leads to a family of gaits:
When the twisting is uniform along the whole body, the entire robot rolls like a wheel during sidewinding, which we call \textit{rolling sidewinding}.
When the twisting varies along the body, different parts of the robot roll at different speeds, which turns the robot, and we call this gait \textit{twisting sidewinding}.
In twisting sidewinding, the point where the rolling speed is zero acts as the turning center, and it can be placed anywhere along the body.

Since the twisting of the LTS is spread over the entire body, all twist joints share the load.
On our hardware implementation of the \abbrTSnake, we compare the two gaits under similar angular velocities and similar total output power during in-place turning.
Under this fair comparison, the two most-loaded twist joints of the existing rolling vector approach carry $96.8\%$ of the total power, while those of our LTS carry only $45.1\%$ (Fig.~\ref{fig:head}(b)).
The LTS reduces the peak output power of a single twist joint by $42\%$, leaving a larger margin before a joint motor saturates (Fig.~\ref{fig:head}(c)).
Besides, our LTS with evenly distributed workload enables the robot to locomote on various complex terrains.
We compare our LTS against conventional sidewinding without twisting on four types of outdoor terrains, and our LTS moves the robot faster than conventional sidewinding, reducing the traversal time by up to $49\%$ for a fixed travel length (Fig.~\ref{fig:head}(d)).
Real-robot experiments also validate the translating and turning motions of the proposed gaits.



\section{Related Works}\label{sec:related_works}
The large number of redundant DoFs makes it challenging to design gaits for snake robots.
Existing methods for snake robot gait design can be classified into central pattern generators~\cite{ijspeert2007swimming,wang2017cpg}, reinforcement learning~\cite{liu2023reinforcement,sheng2024maneuverable,bing2022simulation}, parameterized gait equations~\cite{hatton2010sidewinding,tesch2009parameterized}, and the backbone curve methods~\cite{chirikjian2002modal,zhen2015modeling,tanaka2018development}.
In this paper, we leverage the geometric intuition of the backbone curve method to design and analyze linear twisting sidewinding gaits for the \abbrTSnake.

The proposed gaits are based on the sidewinding locomotion, which is an efficient gait family for snake robots to locomote on complex terrains~\cite{tanev2005automated,melo2015modular,ariizumi2017dynamic,malayjerdi2019analytical}.
Sidewinding achieves both translational and turning maneuvers in many ways.
For example, conical sidewinding enables fixed-point turning~\cite{tanev2005automated,gong2013extended}, and sidewinding with the contact length control technique allows navigation on sandy slopes~\cite{marvi2014sidewinding}.
Recent works \cite{chang2019evaluation,chong2021frequency} further improved sidewinding motions by adding artificial scales on the snake robots and modulating the spatial frequency of the vertical body wave.

Snake robots that can not only bend but also twist are designed and realized in several ways.
The first class adds the limited twisting capability to snake robots' inner backbone, using twist joints~\cite{wang2017gait,zhou2025direct,wang2025u}, cable-driven design~\cite{tanaka2023twist,incekara2025tendon}, and the helical spring backbone~\cite{zhao2021multigait}. 
The twisting angular range of this class of robots is limited.
Another class adds unlimited torsion to outer layers, such as the ARCSnake~\cite{richter2021arcsnake} and EELS~\cite{vaquero2024eels}.
The inner backbones of this class of snake robots still cannot twist unlimitedly.
Different from these existing snake robots, the \abbrTSnake mechanism considered in this paper can twist its backbone unlimitedly and meets the mechanical requirement of the proposed gaits.

\section{Method}\label{sec:method}

This section presents the development of linear twisting sidewinding gaits using the backbone curve method.
We first review the key concepts of the dorsal reference curve approach, which allows flexible design of curve twist (Sec.~\ref{subsec:drc}). 
Adopting the backbone curve of conventional sidewinding gait (Sec.~\ref{subsec:bc_of_sw}), we then describe the construction of linear twisting sidewinding gaits (Sec.~\ref{subsec:bc_of_lts}). 
Finally, to obtain the joint angles required to execute the proposed gaits on the \abbrTSnake, we introduce its joint arrangement and propose a piecewise-approximation integration method based on this arrangement (Sec.~\ref{subsec:from_curve_to_gaits}).

\subsection{Dorsal Reference Curves}\label{subsec:drc}

To represent the geometric shape and axial twist of the \abbrTSnake, we use the dorsal reference curve~\cite{yamada2006study} to describe both properties.
We then review some key concepts in this method, which are used in this paper.

\begin{figure}[tb!]
    \centering
    \includegraphics[width=1.0\linewidth]{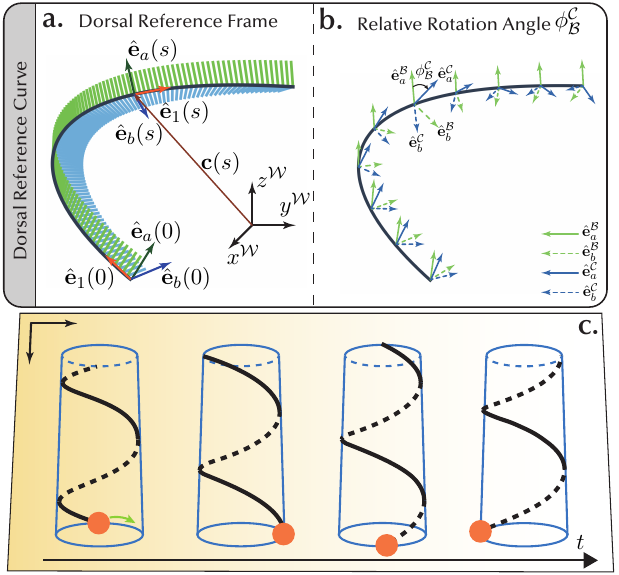}
    \caption{Overview of the dorsal reference curve and the backbone curve of sidewinding. (a) The dorsal reference frames along the curve $\mathbf{c}(s)$. (b) The relative rotation angle $\phi^\mathcal{C}_\mathcal{B}$ between frames $\mathcal{B}$ and $\mathcal{C}$. (c) The backbone curve of sidewinding is a time-varying elliptical helix.}
    \label{fig:dorsal_curve}
\end{figure}

\subsubsection{Dorsal Reference Frames}

As shown in Fig.~\ref{fig:dorsal_curve}(a), let $\mathcal{W}:\{x^\mathcal{W},y^\mathcal{W},z^\mathcal{W}\}$ denote the world frame and $\mathbf{c}(s)=[x(s),y(s),z(s)]^\top\in\mathcal{W}$ denote a smooth spatial curve defined in this frame.
The dorsal reference frames $\{\hat{\mathbf{e}}_1,\hat{\mathbf{e}}_a,\hat{\mathbf{e}}_b\}$ are defined along the curve with three orthonormal vectors: the unit tangent $\hat{\mathbf{e}}_1$ and the normal and binormal vectors $\hat{\mathbf{e}}_a$ and $\hat{\mathbf{e}}_b$, which specify the dorsal-ventral distinction of the curve.
These vectors satisfy the \textit{shape equations}~\cite{yamada2006study}:
\begin{equation}\label{eqn:shape_equation}
    \begin{aligned}
        \frac{{\rm d}} {{\rm d} s}
        \begin{bmatrix}
            \hat{\mathbf{e}}_1 \\ 
            \hat{\mathbf{e}}_a \\
            \hat{\mathbf{e}}_b
        \end{bmatrix}
        = 
        \begin{bmatrix}
            0 & \kappa_b(s) & -\kappa_a(s) \\
            -\kappa_b(s) & 0 & \tau(s) \\
            \kappa_a(s) & -\tau(s) & 0
        \end{bmatrix} 
        \begin{bmatrix}
            \hat{\mathbf{e}}_1 \\ 
            \hat{\mathbf{e}}_a \\
            \hat{\mathbf{e}}_b
        \end{bmatrix}
    \end{aligned}
\end{equation}
where $\hat{\mathbf{e}}_1=\frac{{\rm d}\mathbf{c}}{{\rm d}s}$ and $\kappa_a$, $\kappa_b$, and $\tau$ are the \textit{shape functions}.
The first two describe curvatures along the axes perpendicular to the tangent, while $\tau$ describes torsion along the curve.

\subsubsection{Torsion Description}

Two special cases of dorsal reference curves are used in this paper, namely \textit{bellows model} and \textit{complete model}.
When $\tau=0$ along the entire curve, the \textit{bellows model} describes a robot that can only bend but cannot twist.
This model defines a set of bellows frames along the curve.
To preserve zero torsion, rotating one frame about its tangent requires all other frames to rotate by the same angle. 
In the complete frame, all three shape functions $\kappa_a$, $\kappa_b$, and $\tau$ may be non-zero, so its complete frames describe both bending and twisting along the curve.
We use bellows and complete frames to model rolling and twisting motion, respectively.

At each point $\mathbf{c}(s)$, consider a bellows frame $\mathcal{B}:\{\hat{\mathbf{e}}_1^\mathcal{B},\hat{\mathbf{e}}_a^\mathcal{B},\hat{\mathbf{e}}_b^\mathcal{B}\}$ and a complete frame $\mathcal{C}:\{\hat{\mathbf{e}}_1^\mathcal{C},\hat{\mathbf{e}}_a^\mathcal{C},\hat{\mathbf{e}}_b^\mathcal{C}\}$ obtained by twisting.
They share the tangent vector $\hat{\mathbf{e}}_1^\mathcal{B}=\hat{\mathbf{e}}_1^\mathcal{C}$ and differ by a rotation $\phi_\mathcal{B}^\mathcal{C}(s)$ about this axis from $\mathcal{B}$ to $\mathcal{C}$ (Fig.~\ref{fig:dorsal_curve}(b)).
The spatial variation of this angle describes torsion through the following shape-function relationships~\cite{yamada2006study}:
\begin{subequations}
\label{eqn:shape_functions_transform}
    \begin{align} 
        \begin{bmatrix} 
            \kappa_{a}^\mathcal{C} \\ 
            \kappa_{b}^\mathcal{C} 
        \end{bmatrix} &= 
        \begin{bmatrix} 
            \cos \phi^\mathcal{C}_\mathcal{B} & \sin \phi^\mathcal{C}_\mathcal{B} \\ 
            -\sin \phi^\mathcal{C}_\mathcal{B} & \cos \phi^\mathcal{C}_\mathcal{B} 
        \end{bmatrix} 
        \begin{bmatrix} 
            \kappa_{a}^\mathcal{B} \\ 
            \kappa_{b}^\mathcal{B} 
        \end{bmatrix} \label{eqn:kappas_relationship} \\
        \tau^\mathcal{C} &= \frac{ {\rm d} \phi^\mathcal{C}_\mathcal{B}}{{\rm d} s} + \tau^\mathcal{B},
        \label{eqn:tau_relationships}
    \end{align}
\end{subequations}
which we use to design the proposed gaits.

\subsection{Backbone Curve of Sidewinding}\label{subsec:bc_of_sw}

To model the backbone curve of sidewinding gaits, we adopt the elliptical helix model (Fig.~\ref{fig:dorsal_curve}(c)) analyzed in \cite{hatton2010sidewinding}.
The backbone curve of sidewinding is defined by the following time-varying spatial curve in the world frame:
\begin{equation}
    \label{eqn:parametric_equation}
    \mathbf{c}^\text{eh}(\theta,t)=
    \begin{bmatrix}
        k_{\theta}\theta \\
        k_b\sin{(\theta + \omega_t t)}\\
        k_a\cos{(\theta + \omega_t t)}
    \end{bmatrix}, \quad \theta \in [0,\Theta]
\end{equation}
where $\theta$ is an auxiliary parameter, $t$ is the time, $\omega_t$ is the temporal wave frequency, $k_\theta$ determines the pitch, and $k_b$ and $k_a$ are the horizontal and vertical amplitudes, respectively.

Because the remaining derivation uses arc-length $s$, we convert the parameterization using
\begin{equation}
    \label{eqn:theta_to_arc_length}
    s(\theta,t) = \int_0^\theta \|\frac{\partial \mathbf{c}^\text{eh}(\alpha,t)}{\partial \alpha}\| \,  {\rm d} \alpha,
\end{equation}
where the inverse mapping $\theta(s,t)$ can be obtained numerically. 
Substituting this inverse mapping into Eq.~\eqref{eqn:parametric_equation} gives the arc-length parameterization $\mathbf{c}^\text{eh}(s,t), s \in [0,L]$ used below.


\subsection{Backbone Curve of Linear Twisting Sidewinding}\label{subsec:bc_of_lts}

The proposed gait family combines twisting and sidewinding in a unified way.
First, the \abbrTSnake body is represented using the backbone curve of sidewinding $\mathbf{c}^\text{eh}$.
Then, using the complete model, we define the snake head orientation, derive the bellows-frame shape functions, and apply a linear twist function to obtain the complete frames.
We also show that specific parameter settings in the twist function produce two new gaits: rolling sidewinding and twisting sidewinding.

\begin{figure}[tb]
    \centering
    \includegraphics[width=0.6\linewidth]{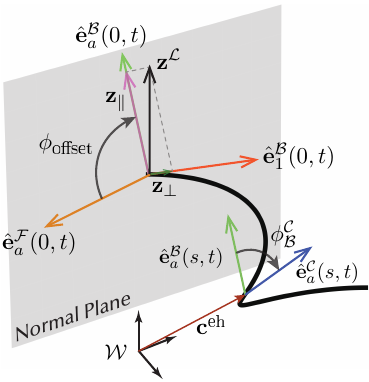}
    \caption{Local frame at snake head. $\mathbf{z}_\parallel$ is obtained by finding the component of $\mathbf{z}^\mathcal{L}$ within the normal plane at $s=0$. The normal vector of the bellows frame on the snake head, $\mathbf{\hat{e}}^\mathcal{B}_a(0,t)$, is computed by normalizing $\mathbf{z}_\parallel$. $\phi_\text{offset}$ defines the angle from $\mathbf{\hat{e}}^\mathcal{F}_a(0,t)$ to $\mathbf{\hat{e}}^\mathcal{B}_a(0,t)$. The twist function $\phi^\mathcal{C}_\mathcal{B}$ defines the rotation angle from the bellows frame $\mathcal{B}$ to the complete frame $\mathcal{C}$.}
    \label{fig:snake_head_ori}
\end{figure}

\subsubsection{Orientation of Snake Head}

Recall that $\mathbf{c}^\text{eh}(s), s\in[0,L]$ is the backbone curve of the robot represented in the world frame $\mathcal{W}$.
For the bellows frame along the curve, let $\mathcal{B}(s,t):\{ \hat{\mathbf{e}}^\mathcal{B}_1(s,t), \hat{\mathbf{e}}^\mathcal{B}_a(s,t), \hat{\mathbf{e}}^\mathcal{B}_b(s,t) \}$ denote the bellows frame at point $s$ along the curve at time $t$. 
Consider the one $\mathcal{B}(0,t)$ that is attached to the snake head, i.e., the beginning of the curve ($s=0$).
The vector $\hat{\mathbf{e}}^\mathcal{B}_1(0,t)$ is aligned with the tangent vector along the curve $\mathbf{c}^\text{eh}(s)$ at $s=0$, while the other two vectors $\hat{\mathbf{e}}^\mathcal{B}_a(0,t), \hat{\mathbf{e}}^\mathcal{B}_b(0,t)$ remain undefined.

To define them, we first introduce a local frame $\mathcal{L}:\{\mathbf{x^\mathcal{L}},\mathbf{y^\mathcal{L}},\mathbf{z^\mathcal{L}}\}$ which translates the origin of the world frame to the start of the curve, while the $x,y,z$-axes of $\mathcal{L}$ remain parallel to those of the world frame (see Fig.~\ref{fig:snake_head_ori}).
We then project $\mathbf{z}^\mathcal{L}$ onto the normal plane at the start of the curve to obtain its component within this plane, $\mathbf{z}_\parallel$. Specifically,
\begin{equation}\label{eqn:snake_head_proj}
    \mathbf{z}_\parallel=\mathbf{z^\mathcal{L}}-\mathbf{z}_\perp,\qquad
    \mathbf{z}_\perp=[\mathbf{z^\mathcal{L}}\cdot\hat{\mathbf{e}}^\mathcal{B}_1(0,t)]\hat{\mathbf{e}}^\mathcal{B}_1(0,t).
\end{equation}
where $\mathbf{z}_\perp$ is obtained by projecting $\mathbf{z^\mathcal{L}}=[0,0,1]^\top$ onto the tangent vector at $s=0$, $\hat{\mathbf{e}}^\mathcal{B}_1(0,t)$.
Subsequently, this vector $\mathbf{z}_\parallel$ is normalized to yield $\hat{\mathbf{e}}^\mathcal{B}_a(0,t)$.
At this point, the bellows frame at the beginning of the curve is well-defined.

\subsubsection{Construction of the Bellows Frame Along the Curve}

To derive the bellows frames along the curve, we compute the shape functions of the curve.
Directly deriving the expressions in the bellows frame can be complicated, and we choose to leverage the Frenet-Serret (F-S) frame to indirectly derive the shape functions, which consists of three steps:
\begin{enumerate}
    \item Construct the Frenet-Serret frame $\mathcal{F}$ along the curve and calculate the curvature $ \kappa^\mathcal{F}$ and torsion $ \tau^\mathcal{F}$ of the curve.
    \item Derive the relative rotation angle $\phi^\mathcal{B}_\mathcal{F}$ from the F-S frame $\mathcal{F}$ to the bellows frame $\mathcal{B}$.
    \item Substitute $ \kappa^\mathcal{F}$, $ \tau^\mathcal{F}$ and $\phi^\mathcal{B}_\mathcal{F}$ into Eq.~\eqref{eqn:kappas_relationship} to derive the curvatures within bellows frame $\mathcal{B}$.
\end{enumerate}

To construct the Frenet-Serret frame $\mathcal{F}:\{ \hat{\mathbf{e}}^\mathcal{F}_1, \hat{\mathbf{e}}^\mathcal{F}_a, \hat{\mathbf{e}}^\mathcal{F}_b \}$ along the curve, according to \cite{do2016differential}, the corresponding curvature $ \kappa^\mathcal{F}$ and torsion $ \tau^\mathcal{F}$ can be calculated as follows:
\begin{subequations}
\label{eqn:geometric shape functions}
    \begin{align}
        \kappa^\mathcal{F}(s,t) &= \| \frac{\partial ^2 \mathbf{c}^\text{eh}}{\partial s^2} \|   \label{eqn:kappa_formula} \\
        \tau^\mathcal{F}(s,t) &= \frac{(\frac{\partial  \mathbf{c}^\text{eh}}{\partial s} \times \frac{\partial ^2 \mathbf{c}^\text{eh}}{\partial s^2}) \cdot \frac{\partial ^3 \mathbf{c}^\text{eh}}{\partial s^3}}{\| \frac{\partial ^2 \mathbf{c}^\text{eh}}{\partial s^2} \|^2}    \label{eqn:tau_formula}
    \end{align}
\end{subequations}

In Eq.~\eqref{eqn:tau_formula}, ``$\times$" denotes the cross product of the vectors. 
At this point, we have obtained the curvature and torsion of the curve within frame $\mathcal{F}$.

The next step is to calculate the relative rotation angle $\phi^\mathcal{B}_\mathcal{F}$ from the F-S frame $\mathcal{F}$ to the bellows frame $\mathcal{B}$. This angle can be derived by applying the following two conditions:
\begin{itemize}
    \item The bellows frame is torsion-free, which means that the torsion within frame $\mathcal{B}$ remains zero, i.e., $\tau^\mathcal{B} = 0$.
    \item The orientation of frame $\mathcal{B}$ has been determined by $\hat{\mathbf{e}}^\mathcal{B}_a(0,t)$. 
\end{itemize}

Note that Eq.~\eqref{eqn:kappas_relationship} is general for any two frames along the curve with a relative rotation between them, and we apply Eq.~\eqref{eqn:kappas_relationship} to the $\mathcal{F}$ and $\mathcal{B}$ frames here.
Employing the first condition, we substitute $\tau^\mathcal{F}$ and $\phi^\mathcal{B}_\mathcal{F}$ into Eq.~\eqref{eqn:tau_relationships} and yield:
\begin{equation}
    \label{eqn:relationship_bt_two_tau}
    \tau^\mathcal{B} = \frac{\partial \phi_\mathcal{F}^\mathcal{B}(s,t)}{\partial s}+\tau^\mathcal{F} =0 .
\end{equation}
According to Eq.~\eqref{eqn:relationship_bt_two_tau}, the angle $\phi_\mathcal{F}^\mathcal{B}$ can be solved by integrating the above equation:
\begin{equation}
    \label{eqn:solution_of_phi_F_B'}
    \phi_\mathcal{F}^\mathcal{B}(s,t)=-\int_0^s \tau^\mathcal{F} (\beta,t) \, {\rm d} \beta + \phi_\text{offset}(t)
\end{equation}
where $\phi_\text{offset}$ is a spatially invariant angle and remains unknown.
In Eq.~\eqref{eqn:solution_of_phi_F_B'}, the integral term compensates for the torsion of the frame $\mathcal{F}$. 
The integral term alone can only rotate the frame $\mathcal{F}$ to a torsion-free state, and cannot ensure that the orientation of frame $\mathcal{F}$ is aligned with frame $\mathcal{B}$. 
This alignment should then be achieved by the spatially invariant angle $\phi_\text{offset}$. 
Recalling that the orientation of frame $\mathcal{B}$ is defined by $\hat{\mathbf{e}}^\mathcal{B}_a(0,t)$, 
we can determine $\phi_\text{offset}$ by calculating the angle from $\hat{\mathbf{e}}^\mathcal{F}_a(0,t)$ to $\hat{\mathbf{e}}^\mathcal{B}_a(0,t)$:
\begin{equation}
    \label{eqn:phi_offset}
    \phi_\text{offset}(t)=\measuredangle{\left(\hat{\mathbf{e}}^\mathcal{F}_a(0,t),\hat{\mathbf{e}}^\mathcal{B}_a(0,t)\right)}
\end{equation}

In Eq.~\eqref{eqn:phi_offset}, $\hat{\mathbf{e}}^\mathcal{B}_a(0,t)$ is obtained by normalizing $\mathbf{z_\parallel}$, while $\hat{\mathbf{e}}^\mathcal{F}_a(0,t)$ can be calculated as follows~\cite{do2016differential}:
\begin{equation}
    \label{eqn:normal_vec_at_0}
     \hat{\mathbf{e}}^\mathcal{F}_a(0,t)=\left. \frac{{\partial ^2 \mathbf{c}^\text{eh}}/{\partial s^2}}{\|{\partial ^2 \mathbf{c}^\text{eh}}/{\partial s^2}\|}\right|_{s=0}
\end{equation}

Then, we substitute the obtained $\kappa^\mathcal{F}$, $\tau^\mathcal{F}$ and $ \phi_\mathcal{F}^\mathcal{B}$ into Eq.~\eqref{eqn:kappas_relationship} to calculate the curvatures $\kappa_a^\mathcal{B}$ and $\kappa_b^\mathcal{B}$ of the curve within the bellows frame $\mathcal{B}$:
\begin{equation}
    \label{eqn:kappas_of_B}
    \begin{bmatrix} 
        \kappa^\mathcal{B}_{a} \\
        \kappa^\mathcal{B}_{b} 
    \end{bmatrix} = 
    \begin{bmatrix} 
        \cos \phi^\mathcal{B}_\mathcal{F} & \sin \phi^\mathcal{B}_\mathcal{F} 
        \\ -\sin \phi^\mathcal{B}_\mathcal{F} & \cos \phi^\mathcal{B}_\mathcal{F} 
    \end{bmatrix} 
    \begin{bmatrix} 
        0 \\ 
        \kappa^\mathcal{F} 
    \end{bmatrix}
\end{equation}

\subsubsection{Linear Twisting and Complete-Frame Shape Functions}

Based on the derived bellows frames along the curve, we can now apply a linear twist function to the bellows frame to obtain the complete frames $\mathcal{C}:\{ \hat{\mathbf{e}}^\mathcal{C}_1, \hat{\mathbf{e}}^\mathcal{C}_a, \hat{\mathbf{e}}^\mathcal{C}_b \}$ along the curve.
These complete frames $\mathcal{C}$ define the actual shape of the \abbrTSnake and drive the motion of the robot.

We first propose the linear twist function that specifies the relative rotation from $\mathcal{B}$ to $\mathcal{C}$ along the curve:
\begin{equation}
    \label{eqn:Phi_B_to_C}
    \phi_\mathcal{B}^\mathcal{C}(s,t) = (a_1 s+a_0)t.
\end{equation}
This simple map is linear in arc-length $s$ and generates motion as time advances.
We call $a_1$ and $a_0$ the \textit{twisting coefficient} and \textit{rolling coefficient}, respectively. 
Modulating them produces a new gait family, \textit{linear twisting gaits}.\footnote{Because the twist function is independent of the backbone curve, it can potentially be combined with arbitrary gait curves. In this paper, we focus on the sidewinding gait.}

Specifically, based on the shape functions in the bellows frames derived in Eq.~\eqref{eqn:kappas_of_B} and the relationship between the bellows frame and the complete frames in Eqs.~\eqref{eqn:shape_functions_transform}, the shape functions of the complete frames are:
\begin{subequations}
\label{eqn:shape_functions_in_Frame_C}
    \begin{align}
        \kappa^\mathcal{C}_{a}(s,t) &= \kappa^\mathcal{F}\sin{(\phi^\mathcal{C}_\mathcal{B}+\phi^\mathcal{B}_\mathcal{F})} \notag \\ 
        &= \kappa^\mathcal{F}\sin{[(a_1 s +a_0) t+\phi^\mathcal{B}_\mathcal{F}]} \label{eqn:kappa_a_of_C_general} \\
         \kappa^\mathcal{C}_{b} (s,t) &= \kappa^\mathcal{F}\cos{[(a_1 s +a_0) t+\phi^\mathcal{B}_\mathcal{F}]}  \label{eqn:kappa_b_of_C_general} \\
        \tau^\mathcal{C}(s,t)& =\frac{\partial \phi^\mathcal{C}_\mathcal{B}}{\partial s} + \tau^\mathcal{B} = a_1 t\label{eqn:tau_of_C_general} 
    \end{align}
\end{subequations}
As shown in Sec.~\ref{subsec:from_curve_to_gaits}, by numerically integrating these shape functions, we can obtain the joint angles of the snake robot and execute the gait.

\begin{figure}[tb!]
    \centering
    \includegraphics[width=1\linewidth]{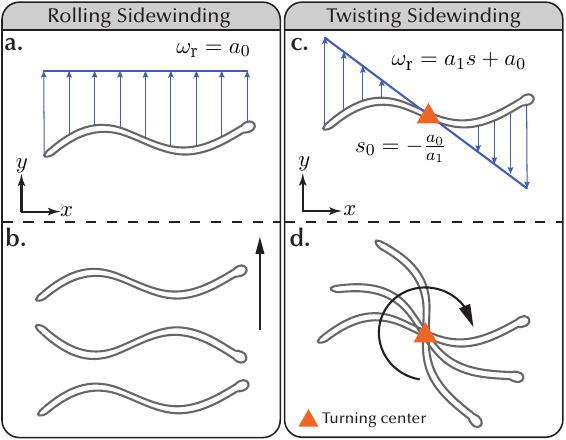}
    \caption{Overview of rolling sidewinding and twisting sidewinding. For rolling sidewinding, $\omega_\text{r}$ (a) is a non-zero constant, providing translating capability (b) for the \abbrTSnake.
    For twisting sidewinding, $\omega_\text{r}$ (c) varies with arc-length linearly, inducing turning motions (d) for the \abbrTSnake.}
    \label{fig:diff_gaits}
\end{figure}

\subsubsection{Gait Specialization}

We show that different parameter settings in the twist function can produce gaits with different locomotion patterns.
The instantaneous angular velocity of the complete frame relative to the bellows frame, which we refer to as the \textit{rolling velocity}, is obtained by differentiating the twist function with respect to time:
\begin{equation}
    \label{eqn:rolling_velocity}
    \omega_{\rm r}(s) = \frac{\partial \phi_\mathcal{B}^\mathcal{C}(s,t)}{\partial t}=a_1s+a_0.
\end{equation}
Thus, $a_1$ determines the spatial gradient of the rolling velocity, whereas $a_0$ determines its value at the head of the backbone curve.

When $a_1=0$, the rolling velocity is uniform along the backbone, $\omega_{\rm r}(s)=a_0$ (Fig.~\ref{fig:diff_gaits}(a)). 
For $a_0\neq 0$, this setting produces \textit{rolling sidewinding}, in which the body rotates about its backbone with a spatially constant angular velocity (Fig.~\ref{fig:diff_gaits}(b)). 
This rolling motion of the \abbrTSnake is conceptually similar to a wheel revolving around its axle.
The limiting case $a_1=a_0=0$ recovers conventional sidewinding without axial rolling.

When $a_1\neq 0$, the rolling velocity varies linearly along the backbone and the resulting gait is \textit{twisting sidewinding} (Fig.~\ref{fig:diff_gaits}(c)). 
Its turning center $s_0$ is the point at which the local rolling velocity vanishes, and is therefore given by
\begin{equation}
    \label{eqn:turning_center}
    \omega_{\rm r}(s_0)=0,\qquad s_0=-\frac{a_0}{a_1}.
\end{equation}
Consequently, by modulating $a_1$ and $a_0$, as shown in Fig.~\ref{fig:diff_gaits}(d), we can design the \abbrTSnake's turning behavior by locating the turning center.

\subsection{From Curve to Gaits}\label{subsec:from_curve_to_gaits}

We describe how to convert the continuous shape functions into joint commands for the discrete \abbrTSnake mechanism. 
We first review the conventional integration method~\cite{yamada2010approximations} and then introduce a piecewise-approximation method for the proposed linear twisting sidewinding gaits.

\subsubsection{Integration for \abbrTSnake}

\begin{figure}[tb!]
    \centering
    \includegraphics[width=1\linewidth]{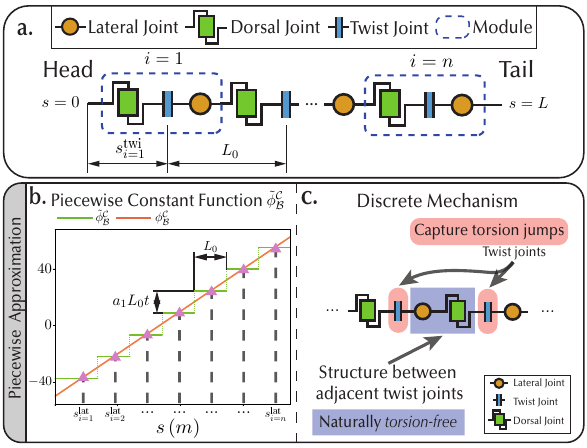}
    \caption{Analysis of the piecewise approximation solution. (a) Illustration of the discrete mechanism of the \abbrTSnake. (b) After the piecewise approximation, the relative rotation angle $\tilde{\phi}_\mathcal{B}^\mathcal{C}$ becomes a piecewise constant function. (c) The correspondence to the discrete \abbrTSnake.} 
    \label{fig:piecewise_approx}
\end{figure}

As shown in Fig.~\ref{fig:piecewise_approx}(a), the \abbrTSnake consists of $n$ identical modules, each of length $L_0$ and containing three 1-DoF revolute joints in a \textit{dorsal-twist-lateral} sequence. 
The dorsal and lateral joint axes pass through the module's cross-sectional center, span the backbone's normal plane, and are mutually orthogonal in the straight configuration; the twist axis is aligned with the local backbone tangent and is therefore orthogonal to both bending axes. 
The complete mechanism has $3n$ DoFs and total length $L=nL_0$. 
These joints provide a discrete realization of the dorsal reference curve: lateral and dorsal rotations represent the accumulated curvatures $\kappa_a$ and $\kappa_b$, respectively, while twist rotations represent the accumulated torsion $\tau$.

The integration approach from~\cite{yamada2010approximations} can be directly applied to the \abbrTSnake by adding an additional integration for the torsion~\cite{fan2026motion}.
For twisting sidewinding gaits ($a_1 \neq 0$), the torsional deformation accumulates within the \abbrTSnake body, inducing turning motions.
By applying the existing integration method for these motions of the \abbrTSnake, we observe that as the torsional deformation increases, the resulting joint angles of the dorsal and lateral joints along the body become numerically unstable, which further leads to distorted motion of the \abbrTSnake and unexpected motion trajectories.
The reason for this numerical problem is left for future work.

\subsubsection{Piecewise-Approximation Integration Method}

We avoid this numerical problem by using a piecewise approximation integration method, which aligns with the discrete mechanism of \abbrTSnake.
For the actual \abbrTSnake mechanism, each link is a rigid body and must be torsion-free, and the torsion can only be achieved at the twist joints (Fig.~\ref{fig:piecewise_approx}(c)).
Based on this physical constraint, we approximate the twist angle $\phi^\mathcal{C}_\mathcal{B}$ from a linear function into a piecewise constant function with respect to arc-length $s$ (Fig.~\ref{fig:piecewise_approx}(b)).
We take the lateral joints as an example, as the dorsal joints can be handled in a similar way.

Let $\theta_i$ and $s_i$ be the joint angle and position of the $i$-th joint of the same type (lateral, dorsal, or twist) along the \abbrTSnake body.
The $i$-th lateral joint covers the range of arc-length $(s_i^\text{lat}-L_0/2, s_i^\text{lat}+L_0/2]$, whose middle point is $s_i^\text{lat}$:
\begin{equation}
    \label{eqn:Phi BC approx}
    \begin{aligned}
        \tilde{\phi}^\mathcal{C}_\mathcal{B}(s,t) = &(a_1 s_i^{\text{lat}} + a_0) t, 
        \text{for } s \in \left( s_i^{\text{lat}} - \frac{L_0}{2}, s_i^{\text{lat}} + \frac{L_0}{2} \right].
    \end{aligned}
\end{equation}
Next, the lateral and dorsal joint angles calculated by our integration method are given as follows: 
\begin{subequations}
\label{eqn:joint angles refined}
    \begin{align}
        {\theta}_i^\text{lat}(t) &= \int_{s_i^\text{lat}-L_0/2}^{s_i^\text{lat}+L_0/2} \kappa^\mathcal{F}\sin{(\tilde{\phi}^\mathcal{C}_\mathcal{B}+\phi^\mathcal{B}_\mathcal{F})} \, {\rm d} s \notag \\
         &= \int_{s_i^\text{lat}-L_0/2}^{s_i^\text{lat}+L_0/2} \kappa^\mathcal{F}\sin{[(a_1 s_i^{\text{lat}} +a_0) t+\phi^\mathcal{B}_\mathcal{F}]} \, {\rm d} s \label{eqn:lateral_angles_refined} \\
        {\theta}_i^\text{dor}(t)  &= \int_{s_i^\text{dor}-L_0/2}^{s_i^\text{dor}+L_0/2} \kappa^\mathcal{F}\cos{[(a_1 s_i^{\text{dor}} +a_0) t+\phi^\mathcal{B}_\mathcal{F}]} \, {\rm d} s \label{eqn:dorsal_angles_refined} 
    \end{align}
\end{subequations}
The twist joint angles can still be computed using:
\begin{equation}
\label{eqn:twist joint angles}
    \theta_i^\text{twi}(t)   = \int_{s_i^\text{twi}-L_0/2}^{s_i^\text{twi}+L_0/2} a_1 t \, {\rm d} s = a_1 L_0 t.
\end{equation}




\section{Experimental Results}\label{sec:exp}
\begin{table}[tb]
    \centering
    \begin{threeparttable}
    \caption{Common Experiment Parameters}
    \label{tab:exp_parameters_grouped}
    \begin{tabular}{ccl}
        \toprule
        Parameter & Value & Description \\ 
        \midrule
        \addlinespace[0.1em]
        $n$ & $6$  & Number of modules \\ 
        \addlinespace[0.1em]
        $k_a$ & $0.02$ m & Vertical amplitude of $\mathbf{c}^\text{eh}$ \\ 
        \addlinespace[0.1em]
        $k_b$ & $0.08$ m & Horizontal amplitude of $\mathbf{c}^\text{eh}$ \\ 
        \addlinespace[0.1em]
        $k_\theta$ & $(1/2\pi)$ m & Pitch of $\mathbf{c}^\text{eh}$ \\ 
        \addlinespace[0.1em]
        $L_0$ & $0.2055$ m  & Module length \\ 
        \addlinespace[0.1em]
        $\Delta t$ & $0.02$ s  & Time step of joint command\\ 
        \addlinespace[0.1em]
        \bottomrule
    \end{tabular}
    \end{threeparttable}
\end{table}

This section shows the experimental results of the linear twisting sidewinding gaits on the \abbrTSnake.
We first run twisting sidewinding in simulation to validate the proposed piecewise approximation method (Sec.~\ref{subsec:valid}).
Next, we use the \abbrTSnake hardware to compare the twist-joint load distribution with that of the existing rolling vector approach~\cite{fan2026motion}, demonstrating that linear twisting can effectively distribute the twist-joint load evenly (Sec.~\ref{subsec:twist_load}).
We then demonstrate different locomotion modes produced by the proposed gaits (Sec.~\ref{subsec:diff_modes}).
Finally, we test the traversal speed of rolling sidewinding over various complex terrains (Sec.~\ref{subsec:terrains}).

It is worth noting that the twist joints of the \abbrTSnake hardware impose no torsional angle limits.\footnote{Our physical \abbrTSnake implements the mechanism in Fig.~\ref{fig:piecewise_approx}(a) as described in Sec.~\ref{subsec:from_curve_to_gaits}, and the design of this robot is neither the focus nor claimed as a contribution of this paper.}
Therefore, $\theta_i^{\mathrm{twi}}$ in Eq.~\eqref{eqn:twist joint angles} never reaches any upper or lower bound.
In each group of experiments, different twisting and rolling coefficients $a_1,a_0$ and the temporal frequency of the backbone curve $\mathbf{c}^\text{eh}$ are combined to generate distinct locomotion modes.
These parameters are specified in the respective subsections.
The common parameters are listed in Table~\ref{tab:exp_parameters_grouped}.

\begin{figure}[tb!]
    \centering
    \includegraphics[width=1\linewidth]{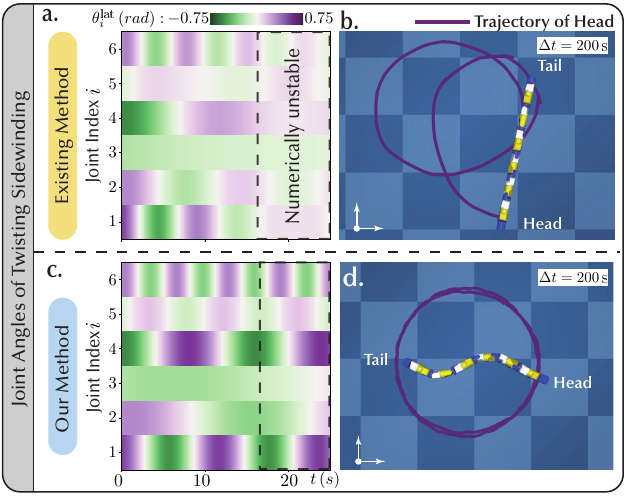}
    \caption{Simulation of the twisting sidewinding. The lateral joint angles computed by the existing method (a) and our method (c). 
    (b) and (d): corresponding snapshots from the simulation experiments.}
    \label{fig:integration}
\end{figure}

\subsection{Validation of the Proposed Integration Method}\label{subsec:valid}

This experiment simulates twisting sidewinding in MuJoCo. 
We execute the joint angles computed by both the existing integration method and our new method (Eqs.~\eqref{eqn:joint angles refined} and \eqref{eqn:twist joint angles}).
We keep $\omega_t$ as zero, $a_1$ as $1.5$ rad/(m$\cdot$s) and $a_0$ as $-1$ rad/s to generate twisting sidewinding. 
We simulate the gaits for $200$ seconds.

We use the lateral joint angles as an example, as the dorsal joint angles exhibit similar behavior.
As shown in Fig.~\ref{fig:integration}(a), the joint angles of twisting sidewinding computed by the existing integration method become numerically unstable as time progresses, causing unexpected locomotion of the \abbrTSnake (Fig.~\ref{fig:integration}(b)).
In contrast, with our method, the computed joint angles are stable and periodic (Fig.~\ref{fig:integration}(c)).
The robot head traces a clear circular path and rotates about the turning center (Fig.~\ref{fig:integration}(d)).
This simulation experiment verifies the existence of the numerical problem and the effectiveness of our integration approach.

\begin{figure}[tb!]
    \centering
    \includegraphics[width=1\linewidth]{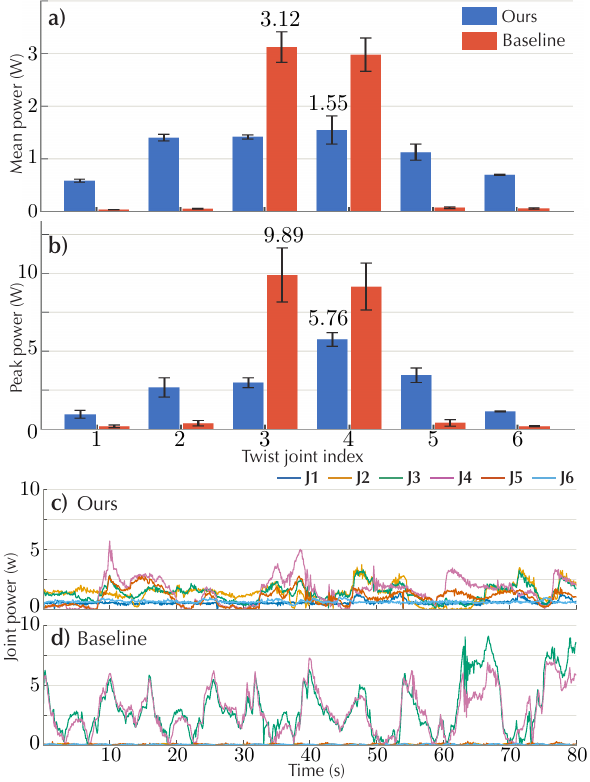}
    \caption{Analysis of twist-joint load distribution. (a) Mean output power of each twist joint. (b) Peak output power of each twist joint. (c) and (d) Instantaneous output power of the six twist joints for the proposed gait and the rolling vector approach, respectively.}
    \label{fig:twist_load}
\end{figure}

\subsection{Twist-Joint Load Distribution}\label{subsec:twist_load}

This experiment compares the twist-joint load distribution between the proposed twisting gait for in-place turning and the rolling vector approach~\cite{fan2026motion}. 
The in-place turning motion is executed on the \abbrTSnake hardware. 
To ensure a fair comparison, we select parameters that make the \abbrTSnake's angular velocities during in-place turning as similar as possible for the two gaits.
For the proposed twisting sidewinding gait, we use $a_0 = 0.6$ rad/s and $a_1 = -2a_0/L$ to perform in-place turning with the turning center at $s_0 = L/2$.
The gait parameters for the rolling vector approach will be provided later. 
We use joint output power to quantify the load on the twist joints. 
This is computed as $p_i= \lvert \tau_i \dot{\theta}^\text{twi}_i \rvert$, where the torque $\tau_i$ is read from the motor current and $\dot{\theta}^\text{twi}_i$ is measured by the joint encoder.
Five trials are conducted for each gait, and the \abbrTSnake steers for $180^\circ$ for each trial.

\subsubsection{Baseline}

The rolling vector approach divides the \abbrTSnake into a head segment, a buffer segment, and a tail segment.
The rolling speed of each segment is mapped by the angular speed of its rolling vector.
By setting these speeds, the \abbrTSnake can achieve a locomotion mode similar to that of a differential-drive robot.
In this experiment, we set the rolling speeds of the head and tail segments, $\omega^\text{h}$ and $\omega^\text{t}$, as $0.4$ and $-0.4$ rad/s, to generate in-place turning motion.

\subsubsection{Results}

Both methods generate similar angular velocities of $2.36 \pm 0.05$ $^\circ$/s (the proposed gait) and $2.33 \pm 0.13$ $^\circ$/s (the rolling vector approach) on the \abbrTSnake.
The total twist-joint output power is also similar, $6.8$ W (the proposed gait) versus $6.3$ W (the rolling vector approach).
The comparison is thus fair. 
The two gaits complete the same task with nearly the same total power.

Despite the similar total power, the two gaits distribute the load very differently (Fig.~\ref{fig:twist_load}(a) and (b)).
For the baseline, $96.8\%$ of the power is carried by the two middle joints (joints 3 and 4), while the other four joints stay almost idle.
For the proposed gait, the power spreads over all six twist joints: the two most-loaded joints carry only $45.1\%$ of the total.
The peak load on a single joint therefore drops from $9.89$ W (joint 3) to
$5.76$ W (joint 4).
The instantaneous power curves in Fig.~\ref{fig:twist_load}(c) and (d) show the same trend over time.

This difference comes from how each gait allocates twisting along the body.
The rolling vector approach confines all twisting to the narrow buffer segment, so the twist joints inside it (joints 3 and 4) bear the whole load.
The proposed gait twists the entire body uniformly, so every twist joint shares the load.
By lowering the peak load on a single joint, the proposed gait keeps more margin before a joint motor saturates, which supports faster turning locomotion.

\begin{figure}[tb!]
    \centering
    \includegraphics[width=1\linewidth]{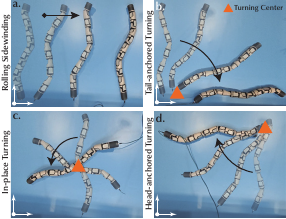}
    \caption{Different locomotion modes of the proposed gaits on the \abbrTSnake.}
    \label{fig:diff_modes}
\end{figure}

\subsection{Different Locomotion Modes}\label{subsec:diff_modes}

In this experiment, we validate the translating and steering capabilities of the proposed gaits.
The experiment is conducted under laboratory conditions, i.e., on hard, flat ground.
We employ the rolling sidewinding ($a_1=0$) for translating and the twisting sidewinding ($a_1 \neq 0$) for steering.

\subsubsection{Translating}

We set $\omega_t=0.5\,\text{rad/s}$, $a_1=0$, and $a_0=0.8\,\text{rad/s}$ to realize this motion.
The resulting locomotion is shown in Fig.~\ref{fig:diff_modes}(a), verifying the effectiveness of the proposed gait on real robots.

\subsubsection{Steering}

By varying the ratio of $a_1$ and $a_0$, we can control the turning centers of twisting sidewinding gaits (Eq.~\eqref{eqn:turning_center}).
We test three modes of steering: $a_0=0.4$ rad/s and $a_1 = -a_0/L$ for tail-anchored turning (Fig.~\ref{fig:diff_modes}(b)), $a_0=-0.4$ rad/s and $a_1 = -2a_0/L$ for in-place turning (Fig.~\ref{fig:diff_modes}(c)), and $a_0=0$ rad/s and $a_1 = -0.8/L$ for head-anchored turning (Fig.~\ref{fig:diff_modes}(d)).
These results demonstrate the steering capability provided by the proposed gait.

\begin{figure}[tb!]
    \centering
    \includegraphics[width=1\linewidth]{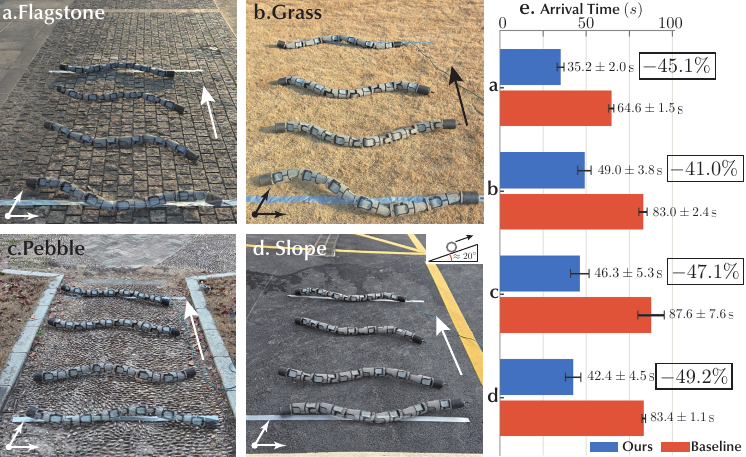}
    \caption{Comparative experiments of traversal speed on four different terrains: (a) flagstone, (b) grass, (c) pebble, (d) slope. The bar chart (e) verifies the proposed gait significantly reduces the arrival time compared to the conventional sidewinding.}
    \label{fig:terrains}
\end{figure}

\subsection{Traversal Speed on Complex Terrains}\label{subsec:terrains}

In this experiment, we show the robustness of the proposed gait on complex terrains.
We compare the traversal speed between our rolling sidewinding and conventional sidewinding~\cite{hatton2010sidewinding} on four types of terrains: flagstone, grass, pebble, and a $20^\circ$ slope.
The experimental setup is illustrated in Fig.~\ref{fig:terrains}.
We use a test track of length $2$ m for all terrains and measure the arrival time.
We fix $\omega_t$ for both gaits at $0.5$ rad/s. 
The conventional sidewinding is configured with $a_0=0$, while the rolling sidewinding uses $a_0=0.8$ rad/s.
Each test is repeated five times on every terrain type.

As shown in Fig.~\ref{fig:terrains}(e), the proposed gait exhibits an obvious reduction in arrival times across all tested terrains compared to the conventional sidewinding. 
Specifically, the time reduction rates are $45.1\%$ on flagstone, $41.0\%$ on grass, $47.1\%$ on pebble, and $49.2\%$ on the slope. 
These results indicate that the rolling sidewinding enhances the traversal speed on various complex terrains.

\section{Conclusion and Future Works}\label{sec:conclude}
In this paper, we propose a novel family of linear twisting sidewinding gaits for twistable snake robots to traverse complex terrains. 
By integrating the twisting motion with the conventional undulatory sidewinding gait, we expand the locomotive capability of the \abbrTSnake. 
Furthermore, to address the numerical problem in the existing integration method, we develop a new approach that leverages the features of the discrete \abbrTSnake mechanism and accurately generates its joint angles. 
Extensive simulation and real-robot experiments demonstrate that the proposed gaits not only lower the peak load on the twist joint when steering, but also achieve faster traversal speeds on various complex terrains.

When computing the joint angles for twisting sidewinding, the reason for the numerical problem in the existing integration method is still underexplored.
Besides, the present study only considers integrating the twist function into the sidewinding gait.
Our future work includes fusing the twist motion with other gaits to better traverse and interact with complex environments.

\section*{Acknowledgments}

OpenAI Codex assisted with plotting code for Fig.~\ref{fig:dorsal_curve}, and with grammar checking and wording revisions.
The authors take responsibility for the final content.

\bibliographystyle{IEEEtran}
\bibliography{references}

@inproceedings{yamada2006study,
  title={Study on the 3D shape of active cord mechanism},
  author={Yamada, Hiroya and Hirose, Shigeo},
  booktitle={Proceedings 2006 IEEE International Conference on Robotics and Automation, 2006. ICRA 2006.},
  pages={2890--2895},
  year={2006},
  organization={IEEE}
}

@inproceedings{hatton2010sidewinding,
  title={Sidewinding on slopes},
  author={Hatton, Ross L and Choset, Howie},
  booktitle={2010 IEEE International Conference on Robotics and Automation},
  pages={691--696},
  year={2010},
  organization={IEEE}
}

@inproceedings{yamada2010approximations,
  title={Approximations to continuous curves of active cord mechanism made of arc-shaped joints or double joints},
  author={Yamada, Hiroya and Hirose, Shigeo},
  booktitle={2010 IEEE international conference on robotics and automation},
  pages={703--708},
  year={2010},
  organization={IEEE}
}

@inproceedings{zhen2015modeling,
  title={Modeling rolling gaits of a snake robot},
  author={Zhen, Weikun and Gong, Chaohui and Choset, Howie},
  booktitle={2015 IEEE international conference on robotics and automation (ICRA)},
  pages={3741--3746},
  year={2015},
  organization={IEEE}
}

@book{do2016differential,
  title={Differential geometry of curves and surfaces: revised and updated second edition},
  author={Do Carmo, Manfredo P},
  year={2016},
  publisher={Courier Dover Publications}
}

@article{zhou2025direct,
  title={A Direct Discretization Method of Backbone Curve for Snake Robots With Multiple Roll Joints},
  author={Zhou, Yunhu and Wang, Shen and Huang, Darong and Ni, Fenglei and Na, Yuhong and Liu, Hong},
  journal={IEEE/ASME Transactions on Mechatronics},
  year={2025},
  publisher={IEEE}
}

@article{vaquero2024eels,
  title={EELS: Autonomous snake-like robot with task and motion planning capabilities for ice world exploration},
  author={Vaquero, Tiago Stegun and Daddi, Guglielmo and Thakker, Rohan and Paton, Michael and Jasour, A and Strub, Marlin P and Swan, R Michael and Royce, Rob and Gildner, Matthew and Tosi, P and others},
  journal={Science robotics},
  volume={9},
  number={88},
  pages={eadh8332},
  year={2024},
  publisher={American Association for the Advancement of Science}
}

@article{marvi2014sidewinding,
  title={Sidewinding with minimal slip: Snake and robot ascent of sandy slopes},
  author={Marvi, Hamidreza and Gong, Chaohui and Gravish, Nick and Astley, Henry and Travers, Matthew and Hatton, Ross L and Mendelson III, Joseph R and Choset, Howie and Hu, David L and Goldman, Daniel I},
  journal={Science},
  volume={346},
  number={6206},
  pages={224--229},
  year={2014},
  publisher={American Association for the Advancement of Science}
}

@article{astley2015modulation,
  title={Modulation of orthogonal body waves enables high maneuverability in sidewinding locomotion},
  author={Astley, Henry C and Gong, Chaohui and Dai, Jin and Travers, Matthew and Serrano, Miguel M and Vela, Patricio A and Choset, Howie and Mendelson III, Joseph R and Hu, David L and Goldman, Daniel I},
  journal={Proceedings of the National Academy of Sciences},
  volume={112},
  number={19},
  pages={6200--6205},
  year={2015},
  publisher={National Academy of Sciences}
}

@inproceedings{gong2013extended,
  title={Extended gait equation for sidewinding},
  author={Gong, Chaohui and Travers, Matthew J and Fu, Xiaozhou and Choset, Howie},
  booktitle={2013 IEEE International Conference on Robotics and Automation},
  pages={5162--5167},
  year={2013},
  organization={IEEE}
}

@article{chong2021frequency,
  title={Frequency modulation of body waves to improve performance of sidewinding robots},
  author={Chong, Baxi and Wang, Tianyu and Rieser, Jennifer M and Lin, Bo and Kaba, Abdul and Blekherman, Grigoriy and Choset, Howie and Goldman, Daniel I},
  journal={The International Journal of Robotics Research},
  volume={40},
  number={12-14},
  pages={1547--1562},
  year={2021},
  publisher={SAGE Publications Sage UK: London, England}
}

@article{tesch2009parameterized,
  title={Parameterized and scripted gaits for modular snake robots},
  author={Tesch, Matthew and Lipkin, Kevin and Brown, Isaac and Hatton, Ross and Peck, Aaron and Rembisz, Justine and Choset, Howie},
  journal={Advanced Robotics},
  volume={23},
  number={9},
  pages={1131--1158},
  year={2009},
  publisher={Taylor \& Francis}
}

@article{pettersen2017snake,
  title={Snake robots},
  author={Pettersen, Kristin Y},
  journal={Annual Reviews in Control},
  volume={44},
  pages={19--44},
  year={2017},
  publisher={Elsevier}
}

@article{wang2017cpg,
  title={CPG-inspired locomotion control for a snake robot basing on nonlinear oscillators},
  author={Wang, Zhelong and Gao, Qin and Zhao, Hongyu},
  journal={Journal of Intelligent \& Robotic Systems},
  volume={85},
  number={2},
  pages={209--227},
  year={2017},
  publisher={Springer}
}

@article{ijspeert2007swimming,
  title={From swimming to walking with a salamander robot driven by a spinal cord model},
  author={Ijspeert, Auke Jan and Crespi, Alessandro and Ryczko, Dimitri and Cabelguen, Jean-Marie},
  journal={science},
  volume={315},
  number={5817},
  pages={1416--1420},
  year={2007},
  publisher={American Association for the Advancement of Science}
}

@article{chirikjian2002modal,
  title={A modal approach to hyper-redundant manipulator kinematics},
  author={Chirikjian, Gregory S and Burdick, Joel W},
  journal={IEEE Transactions on robotics and automation},
  volume={10},
  number={3},
  pages={343--354},
  year={2002},
  publisher={IEEE}
}

@article{tanaka2018development,
  title={Development and control of articulated mobile robot for climbing steep stairs},
  author={Tanaka, Motoyasu and Nakajima, Mizuki and Suzuki, Yosuke and Tanaka, Kazuo},
  journal={IEEE/ASME Transactions on Mechatronics},
  volume={23},
  number={2},
  pages={531--541},
  year={2018},
  publisher={IEEE}
}

@article{wang2017gait,
  title={Gait planning of a scale-driven snake-like robot for winding and obstacle crossing},
  author={Wang, Wei and Yang, Zhiyong and Guo, Lei and Wu, Gongping and Bai, Yucheng and Li, Shuixia},
  journal={Advances in Mechanical Engineering},
  volume={9},
  number={11},
  pages={1687814017731038},
  year={2017},
  publisher={SAGE Publications Sage UK: London, England}
}

@article{richter2021arcsnake,
  title={Arcsnake: Reconfigurable snakelike robot with archimedean screw propulsion for multidomain mobility},
  author={Richter, Florian and Gavrilov, Peter V and Lam, Hoi Man and Degani, Amir and Yip, Michael C},
  journal={IEEE Transactions on Robotics},
  volume={38},
  number={2},
  pages={797--809},
  year={2021},
  publisher={IEEE}
}

@article{zhao2021multigait,
  title={A multigait continuous flexible snake robot for locomotion in complex terrain},
  author={Zhao, Wei and Wang, Jiangbei and Fei, Yanqiong},
  journal={IEEE/ASME Transactions on Mechatronics},
  volume={27},
  number={5},
  pages={3751--3761},
  year={2021},
  publisher={IEEE}
}

@inproceedings{tanaka2023twist,
  title={Twist Snake: Plastic table-top cable-driven robotic arm with all motors located at the base link.},
  author={Tanaka, Kazutoshi and Hamaya, Masashi},
  booktitle={ICRA},
  pages={7345--7351},
  year={2023}
}

@inproceedings{wang2025u,
  title={U-Snake: A Small-Sized Smart Underwater Snake Robot},
  author={Wang, Bowen and Zuo, Haobo and Fu, Changhong},
  booktitle={2025 IEEE/RSJ International Conference on Intelligent Robots and Systems (IROS)},
  pages={16273--16280},
  year={2025},
  organization={IEEE}
}

@article{incekara2025tendon,
  title={Tendon-Driven Compliant Wheel-Less Snake Robot for Undulatory Locomotion Using Conformable Ground Contacts},
  author={Incekara, Serdar and Kwon, Seongil and Kwon, Gangil and Ha, Junhyoung},
  journal={Advanced Intelligent Systems},
  pages={e202501183},
  year={2025},
  publisher={Wiley Online Library}
}

@article{ariizumi2017dynamic,
  title={Dynamic analysis of three snake robot gaits},
  author={Ariizumi, Ryo and Matsuno, Fumitoshi},
  journal={IEEE Transactions on Robotics},
  volume={33},
  number={5},
  pages={1075--1087},
  year={2017},
  publisher={IEEE}
}

@article{malayjerdi2019analytical,
  title={Analytical modeling of a 3-D snake robot based on sidewinding locomotion},
  author={Malayjerdi, Mohsen and Akbarzadeh, Alireza},
  journal={International Journal of Dynamics and Control},
  volume={7},
  number={1},
  pages={83--93},
  year={2019},
  publisher={Springer}
}

@inproceedings{melo2015modular,
  title={Modular snake robot velocity for side-winding gaits},
  author={Melo, Kamilo},
  booktitle={2015 IEEE International Conference on Robotics and Automation (ICRA)},
  pages={3716--3722},
  year={2015},
  organization={IEEE}
}

@article{tanev2005automated,
  title={Automated evolutionary design, robustness, and adaptation of sidewinding locomotion of a simulated snake-like robot},
  author={Tanev, Ivan and Ray, Thomas and Buller, Andrzej},
  journal={IEEE Transactions on robotics},
  volume={21},
  number={4},
  pages={632--645},
  year={2005},
  publisher={IEEE}
}

@article{chang2019evaluation,
  title={Evaluation of bio-inspired scales on locomotion performance of snake-like robots},
  author={Chang, Alexander H and Vela, Patricio A},
  journal={Robotica},
  volume={37},
  number={8},
  pages={1302--1319},
  year={2019},
  publisher={Cambridge University Press}
}

@article{bing2022simulation,
  title={Simulation to real: learning energy-efficient slithering gaits for a snake-like robot},
  author={Bing, Zhenshan and Cheng, Long and Huang, Kai and Knoll, Alois},
  journal={IEEE Robotics \& Automation Magazine},
  volume={29},
  number={4},
  pages={92--103},
  year={2022},
  publisher={IEEE}
}

@article{liu2023reinforcement,
  title={Reinforcement learning of CPG-regulated locomotion controller for a soft snake robot},
  author={Liu, Xuan and Onal, Cagdas D and Fu, Jie},
  journal={IEEE Transactions on Robotics},
  volume={39},
  number={5},
  pages={3382--3401},
  year={2023},
  publisher={IEEE}
}

@article{sheng2024maneuverable,
  title={A Maneuverable Winding Gait for Snake Robots Based on a Delay-Aware Swing and Grasp Framework Combining Rules and Learning Methods (RA-L)},
  author={Sheng, Fengwei and Wan, Fuxi and Tang, Chaoquan and Guo, Xian},
  journal={IEEE Robotics and Automation Letters},
  year={2024},
  publisher={IEEE}
}

@inproceedings{fan2026motion,
  author = {Fan, Qingling and Ren, Zhongqiang},
  booktitle = {2026 IEEE 22nd International Conference on Automation Science and Engineering (CASE)},
  title = {Motion Planning and Locomotive Reduction for Twistable Snake Robots},
  year = {2026}
}

\end{document}